\documentclass[10pt,twocolumn,letterpaper]{article}

\usepackage[pagenumbers]{cvpr} 
\usepackage[dvipsnames]{xcolor}
\usepackage{etoolbox}
\usepackage{longtable}
\usepackage{pdflscape}
\usepackage{graphicx}
\usepackage{siunitx}
\usepackage{amsmath}
\usepackage{amssymb}
\usepackage{nicefrac}
\usepackage{colortbl}%
  \newcommand{\myrowcolour}{\rowcolor[gray]{0.925}}
  
  \newcommand{\highest}[1]{\textcolor{Maroon}{\textbf{#1}}}

\usepackage{booktabs}

\usepackage[dvipsnames]{xcolor}

\newcommand{\ourmethod}{DrivingGaussian}

\newif\ifdrafting
\draftingtrue 
\ifdrafting
    \newcommand{\todo}[1]{{\leavevmode\color[rgb]{1,0,0}[TODO: #1]}}
    
    \definecolor{tabfirst}{rgb}{1, 0.7, 0.7} 
    \definecolor{tabsecond}{rgb}{1, 0.85, 0.7} 
    \definecolor{tabthird}{rgb}{1, 1, 0.7} 
\else
    \newcommand{\todo}[1]{}
    \newcommand{\ds}[1]{}
    \newcommand{\mh}[1]{}
    
\fi

\newcommand{\aftertab}{\vspace{-1em}}
\newcommand{\afterfig}{\vspace{-1.25em}}
\newcommand{\aroundeqn}{\vspace{-.2em}}

\definecolor{cvprblue}{rgb}{0.21,0.49,0.74}
\usepackage[pagebackref,breaklinks,colorlinks,citecolor=cvprblue,linkcolor=cvprblue,urlcolor=cvprblue]{hyperref}

\def\paperID{xxxx} 
\def\confName{xxxx}
\def\confYear{2024}

\title{\ourmethod{}: Composite Gaussian Splatting for Surrounding \\ Dynamic Autonomous Driving Scenes}

\author{
Xiaoyu Zhou\textsuperscript{1}
~ Zhiwei Lin\textsuperscript{1}
~ Xiaojun Shan\textsuperscript{1}
~ Yongtao Wang\textsuperscript{1}
\thanks{Corresponding author.} \\
~ Deqing Sun\textsuperscript{2 \dag}
~ Ming-Hsuan Yang\textsuperscript{2,3} \thanks{Equal contribution.} \\
{\textsuperscript{1}Wangxuan Institute of Computer Technology, Peking University}\\ 
{\textsuperscript{2}Google Research}
~ {\textsuperscript{3}University of California, Merced} \\
}

\begin{document}
\maketitle
\begin{abstract}
We present DrivingGaussian, an efficient and effective framework for surrounding dynamic autonomous driving scenes. For complex scenes with moving objects, we first sequentially and progressively model the static background of the entire scene with incremental static 3D Gaussians. We then leverage a composite dynamic Gaussian graph to handle multiple moving objects, individually reconstructing each object and restoring their accurate positions and occlusion relationships within the scene. We further use a LiDAR prior for Gaussian Splatting to reconstruct scenes with greater details and maintain panoramic consistency. DrivingGaussian outperforms existing methods in dynamic driving scene reconstruction and enables photorealistic surround-view synthesis with high-fidelity and multi-camera consistency.
Our project page is at: \url{https://github.com/VDIGPKU/DrivingGaussian}.
\end{abstract}
\section{Introduction}
\label{sec:intro}

\begin{figure*}[htbp]
  \centering
  \includegraphics[width=\linewidth]{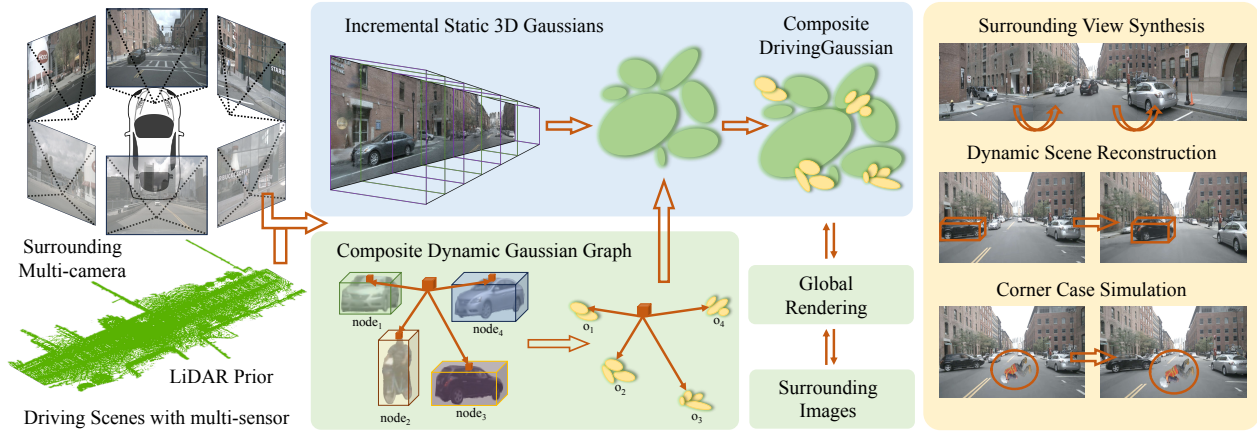}
  \caption{\textbf{Overall pipeline of our method.} \textbf{Left:} \ourmethod{} takes sequential data from multi-sensor, including multi-camera images and LiDAR. \textbf{Middle:} To represent the large-scale dynamic driving scenes, we propose Composite Gaussian Splatting, which consists of two components. The first part incrementally reconstructs the extensive static background, while the second constructs multiple dynamic objects with a Gaussian graph and dynamically integrates them into the scene. \textbf{Right:} \ourmethod{} demonstrates good performance across multiple tasks and application scenarios.}
  \label{fig:overview}
  \afterfig
\end{figure*}

Representing and modeling large-scale dynamic scenes serves as the foundation for 3D scene understanding and contributes to a series of autonomous driving tasks, such as BEV perception~\cite{liu2023bevfusion, li2022bevformer, liang2022bevfusion}, 3D detection~\cite{chen2023futr3d, cai2023objectfusion}, and motion planning~\cite{teng2023motion, caesar2021nuplan}. View synthesis and controllable simulation for driving scenes also enable the generation of corner cases, and safety-critical situations aid in validating and enhancing the safety of autonomous driving systems at a lower cost. 

Unfortunately, reconstructing such complex 3D scenes from sparse vehicle-mounted sensor data is challenging, especially when the ego vehicle moves at high speeds.
Imagine a scene where a vehicle emerges at the edge of an unbounded scene captured by the left-front camera, swiftly moves to the center of the front camera's view, and in the subsequent frames, diminishes into a distant dot.
For such driving scenes, both ego vehicles and dynamic objects are moving at relatively high speeds, posing significant challenges to the scene's construction. The static background and dynamic objects undergo rapid changes, depicted through limited views.
Additionally, it becomes even more challenging in multi-camera settings due to their outward views, minimal overlaps, and variations in light from different directions. 
Complex geometry, diverse optical degradation, and spatiotemporal inconsistency also pose significant challenges to modeling such a 360-degree large-scale driving scene.

Neural radiance fields~\cite{mildenhall2021nerf} (NeRF) has recently emerged as a promising neural reconstruction method for modeling object-level or room-level scenes. Some recent studies~\cite{tancik2022block, turki2022mega, zhenxing2022switch, wang2023neural} have extended NeRF to large-scale, unbounded static scenes, while some focus on modeling multiple dynamic objects within the scene~\cite{ost2021neural, song2022towards}.

However, NeRF-based methods are computationally intensive and require densely overlapping views and consistent lighting. These limit their ability to construct driving scenes with outward multi-camera setups at high speeds. Furthermore, network capacity limitations make it challenging for them to model long-term, dynamic scenes with multiple objects, leading to visual artifacts and blurring.

In contrast to NeRF, the 3D Gaussian Splatting (3D-GS)~\cite{kerbl20233d} represents scenes with more explicit 3D Gaussian representation and achieves impressive performance in novel view synthesis. 
However, the original 3D-GS still encounters significant challenges in modeling large-scale dynamic driving scenes due to fixed Gaussians and constrained representation capacity. Some efforts~\cite{wu20234d, yang2023deformable, luiten2023dynamic} have extended 3D-GS to dynamic scenes by constructing Gaussians at each timestamp. 
Unfortunately, they focus on individual dynamic objects and fail to handle complex driving scenes involving combined static-dynamic regions and multiple moving objects at high speeds.

In this paper, we introduce \ourmethod{}, a novel framework that represents surrounding dynamic autonomous driving scenes.
Our key idea is to model the complex driving scene hierarchically using sequential data from multiple sensors.
We adopt Composite Gaussian Splatting to decompose the whole scene into static background and dynamic objects, reconstructing each part separately. Specifically, we first use incremental static 3D Gaussians to construct comprehensive scenes from surrounding multi-camera views sequentially. We then employ a composite dynamic Gaussian graph to individually reconstruct each moving object and dynamically integrate them into the static background based on the Gaussian graph. Based on these, global rendering via Gaussian Splatting captures occlusion relationships in the real world, encompassing static backgrounds and dynamic objects. Further, we incorporate a LiDAR prior in the GS representation, which is capable of recovering more precise geometry and maintaining better multi-view consistency than utilizing point clouds generated by random initialization or SfM~\cite{schonberger2016structure}.

Extensive experiments demonstrate that our method achieves state-of-the-art performance on public autonomous driving datasets. Even without LiDAR prior, our method still presents promising performance, demonstrating its versatility in reconstructing large-scale dynamic scenes. In addition, our framework enables dynamic scene construction and corner case simulation, which facilitates the validation of the safety and robustness of autonomous driving systems.

\noindent The main contributions of this work are:
\begin{itemize}
\item To our knowledge, \ourmethod{} is the first representation and modeling framework for large-scale, dynamic driving scenes based on Composite Gaussian Splatting. 
\item Two novel modules are introduced, including Incremental Static 3D Gaussians and Composite Dynamic Gaussian Graphs. The former reconstructs the static background incrementally, while the latter models multiple dynamic objects with a Gaussian graph.
Assisted by LiDAR prior, the proposed method facilitates the recovery of complete geometry in large-scale driving scenes.
\item Comprehensive experiments show that DrivingGaussian outperforms previous methods in challenging autonomous driving benchmarks and enables corner case simulation for various downstream tasks.
\end{itemize}
\section{Related Work}
\label{sec:related}

\paragraph{NeRF for Bounded Scenes.}
The rapid progress in neural rendering for novel view synthesis has received significant attention. Neural Radiance Fields (NeRF), which utilizes multi-layer perceptrons (MLPs) and differentiable volume rendering, can reconstruct 3D scenes and synthesize novel views from a set of 2D images and corresponding camera pose information. 
However, NeRF is limited to bounded scenes, requiring a consistent distance between the center object and the camera. It also struggles with scenes captured with slight overlaps and outward capture methods.
Numerous advancements have expanded the capabilities of NeRF, leading to notable improvements in training speed~\cite{muller2022instant, fridovich2022plenoxels, garbin2021fastnerf}, pose optimization~\cite{lin2021barf, wang2021nerf, bian2023nope}, scene editing~\cite{rudnev2022nerf, li2023climatenerf}, and dynamic scene representation~\cite{pumarola2021d, huang2022hdr}.
Nevertheless, applying NeRF to large-scale unbounded scenes, such as autonomous driving scenarios, remains a challenge.

\vspace{-4mm}
\paragraph{NeRF for Unbounded Scenes.}
For large-scale unbounded scenes, ~\cite{tancik2022block, turki2022mega, zhenxing2022switch, wang2023neural, martin2021nerf} have introduced refined versions of NeRF to model multi-scale urban-level static scenes. Inspired by the mipmapping approach to preventing aliasing,~\cite{barron2021mip, barron2022mip} extend NeRF to unbounded scenes. To enable high-fidelity rendering,~\cite{xu2023grid} combines the compact multi-resolution ground feature planes with NeRF for large urban scenes.~\cite{guo2023streetsurf} proposes a close-range-vs-distant-view disentanglement approach, which can model unbounded street views but ignores dynamic objects on the road.
However, these methods model scenes under the assumption that the scene remains static and face challenges in effectively capturing dynamic elements.

Meanwhile, previous NeRF-based methods highly relied on accurate camera poses. Without precise poses,~\cite{liu2023robust, meuleman2023progressively} enable synthesis from dynamic monocular video. However, these methods are confined to forward monocular viewpoints and encounter challenges when dealing with inputs from surrounding multi-camera setups.
For dynamic urban scenes, ~\cite{ost2021neural, song2022towards} extend NeRF to dynamic scenes with multiple objects using a scene graph. ~\cite{wu2023mars, yang2023unisim} propose instance-aware, modular, and realistic simulators for monocular dynamic scenes. ~\cite{xie2023s} improves parameterization and camera poses of surrounding views while using LiDAR as additional depth supervision. ~\cite{turki2023suds, yang2023emernerf} decompose the scene into static background and dynamic objects and constructs the scene with the help of LiDAR and 2D optical flow.

The quality of views synthesized by the aforementioned NeRF-based methods deteriorates in scenarios with multiple dynamic objects and variations and lighting variations, owing to their dependency on ray sampling. In addition, the utilization of LiDAR is confined to providing auxiliary depth supervision, and its potential benefits in reconstruction, such as providing geometric priors, are not explored.

To address these limitations, we utilize Composite Gaussian Splatting to model the unbounded dynamic scenes, where the static background is incrementally reconstructed as the ego vehicle moves, and multiple dynamic objects are modeled and integrated into the entire scene through Gaussian graphs.
LiDAR is employed as the initialization for Gaussians, providing a more accurate geometric shape prior and a comprehensive scene description rather than solely serving as depth supervision for images.

\vspace{-4mm}
\paragraph{3D Gaussian Splatting.}
Recent 3D Gaussian Splatting (3D-GS)~\cite{kerbl20233d} model a static scene with numerous 3D Gaussians, achieving optimal results in novel view synthesis and training speeds. Compared with previous explicit scene representations (e.g., mesh, voxels), the 3D-GS can model complex shapes with fewer parameters. Unlike implicit neural rendering, the 3D-GS allows fast rendering and differentiable computation with splat-based rasterization.

\vspace{-4mm}
\paragraph{Dynamic 3D Gaussian Splatting.}
The original 3D-GS is designed to represent static scenes, and some researchers have extended it to dynamic objects/scenes. Given a set of dynamic monocular images, ~\cite{yang2023deformable} introduces a deformation network to model the motion of Gaussians. ~\cite{wu20234d} connects adjacent Gaussians via a HexPlane, enabling real-time rendering. However, these two approaches are explicitly designed for monocular single-camera scenes focused on a center object. 
\cite{luiten2023dynamic} parameterizes the entire scene using a set of dynamic Gaussians that evolve. However, it requires a camera array with dense multi-view as inputs.

In real-world autonomous driving scenes, the high-speed movement of data collection platforms leads to extensive and complex background variations, often captured by sparse views (e.g., 2-4 views). Moreover, fast-moving dynamic objects with intense spatial changes and occlusion further complicate the situation. Collectively, these factors pose significant challenges for existing methods.
\section{Method}
\label{sec:method}

\subsection{Composite Gaussian Splatting}
3D-GS performs well in purely static scenes but has significant limitations in mixed scenes involving large-scale static backgrounds and multiple dynamic objects. 
As illustrated in Figure~\ref{fig:overview}, we aim to represent surrounding large-scale driving scenes with Composite Gaussian Splatting for unbounded static backgrounds and dynamic objects.

\begin{figure}[t]
  \centering
  \includegraphics[width=\linewidth]{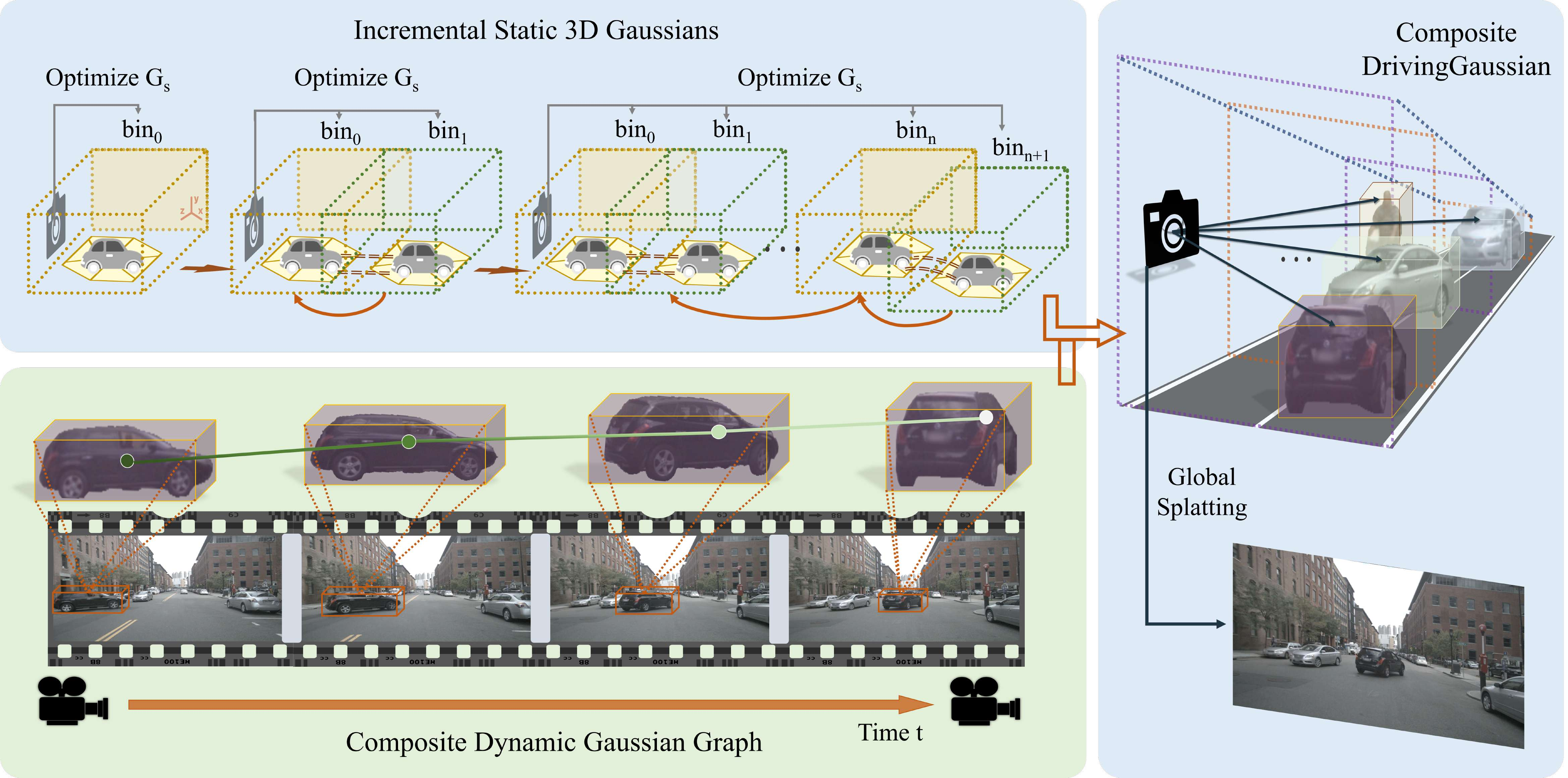}
  \caption{\textbf{Composite Gaussian Splatting with Incremental Static 3D Gaussians and Dynamic Gaussian Graph.} We adopt Composite Gaussian Splatting to decompose the whole scene into static background and dynamic objects, reconstructing each part separately and integrating them for global rendering.}
  \label{static-dynamic}
  \afterfig
 \end{figure}

\vspace{-4mm}
\paragraph{\textbf{Incremental Static 3D Gaussians.}}
\label{static}
The static backgrounds of driving scenes pose challenges due to their large-scale, long-duration, and variations with ego vehicle movement with multi-camera transformation.
As the ego vehicle moves, the static background frequently undergoes temporal shifts and changes. Due to the perspective principle, prematurely incorporating distant street scenes from time steps far away from the current can lead to scale confusion, resulting in unpleasant artifacts and blurring. 
To solve this, we enhance 3D-GS by introducing Incremental Static 3D Gaussians, leveraging the perspective changes introduced by the vehicle's movement and the temporal relationships between adjacent frames, as shown in Figure~\ref{static-dynamic}.

Specifically, we first uniformly divide the static scene into N bins based on the depth range provided by LiDAR prior (Section~\ref{LiDAR}). These bins are arranged in chronological order, denoted as $\{\textrm{b}_i\}^{N}$, where each bin contains multi-camera images from one or more time steps.
For the scene within the first bin, we initialize the Gaussian model using the LiDAR prior (similarly applicable to SfM points):
\aroundeqn
\begin{equation}
  p_{b_0}(l | \mu, \Sigma) = e^{-\frac{1}{2}(l-\mu)^{\top} \Sigma^{-1}(l-\mu)}
\end{equation}
\aroundeqn
where $l \in \mathbb{R}^{3}$ is the position of the LiDAR prior; $\mu$ is the mean of the LiDAR points; $\Sigma \in \mathbb{R}^{3 \times 3}$ is an anisotropic covariance matrix; and $^\top$ is the transpose operator. 
We utilize the surrounding views within this bin segment as supervision to update the parameters of the Gaussian model, including position $P(x,y,z)$, covariance matrix $\Sigma$, coefficients of spherical harmonics for view-dependent color $C(r,g,b)$, along with an opacity $\alpha$.

For the subsequent bins, we use the Gaussians from the previous bin as the position priors and align the adjacent bins based on their overlapping regions. The 3D center for each bin can be defined as:
\aroundeqn
\begin{equation}
  \hat{P}_{b+1}(G_s) = P_{b}(G_s) \bigcup (x_{b+1}, y_{b+1}, z_{b+1})
\end{equation}
\aroundeqn
where $\hat{P}$ is the collection of 3D center for Gaussians $G_s$ of all currently visible regions, $(x_{b+1}, y_{b+1}, z_{b+1})$ is the Gaussians coordinate within the $b+1$ region.
Iteratively, we incorporate scenes from the subsequent bins into the previously constructed Gaussians with multiple surrounding frames as supervision. The incremental static Gaussian model $G_s$ can be defined as:
\begin{equation}
\begin{aligned}
    \hat{C}(G_{s}) = \sum_{b=1}^{N} \Gamma_{b} \ \alpha_{b} \ C_{b}, \quad
     \Gamma_{b} = \prod_{i=1}^{b-1} (1-\alpha_{b})
\end{aligned}
\end{equation}
where $C$ denotes the color corresponding to each single Gaussian at a certain view, $\alpha$ is the opacity, and $\Gamma$ is the accumulated transmittance of the scene according to the $\alpha$ at all the bins. During this process, the overlapping regions between surrounding multi-camera images are used to form the Gaussian models' implicit alignment jointly.

Note that during the incremental construction of static Gaussian models, there might be differences in sampling the same scene between the front and rear cameras. To address this, we employ a weighted averaging to reconstruct the scene's colors as accurately as possible during the 3D Gaussian projection:
\aroundeqn
\begin{equation}
  \tilde{C} = \varsigma(G_{s}) \sum \omega(\hat{C}(G_{s})|R,T)
  \label{eq:color}
\end{equation}
\aroundeqn
where $\tilde{C}$ is the optimized pixel color, $\varsigma$ denotes the differential splatting, 
$\omega$ is the weight for different views, $[R,T]$ is the view-matirx for aligning multi-camera views. 

\vspace{-4mm}
\paragraph{\textbf{Composite Dynamic Gaussian Graph.}}

The autonomous driving environment is highly complex, involving multiple dynamic objects and temporal changes. As shown in Figure~\ref{static-dynamic}, objects are often observed from limited views (e.g., 2-4 views) due to the movements of the ego vehicle and dynamic objects. The high speed also leads to significant spatial changes in dynamic objects, making it challenging to represent them using fixed Gaussians.

To address the challenges, we introduce the Composite Dynamic Gaussian Graph, enabling the construction of multiple dynamic objects in large-scale, long-term driving scenes. 
We first decompose dynamic foreground objects from static backgrounds to build the dynamic Gaussian graph using bounding boxes provided by the datasets. Dynamic objects are identified by their object ID and the corresponding timestamps of appearance. Additionally, the Grounded SAM Models~\cite{ren2024grounded} are employed for precise pixel-wise extraction of dynamic objects based on the range of bounding boxes.

We then build the dynamic Gaussian graph as 
\aroundeqn
\begin{equation} 
H = <O, G_d, M, P, A, T>,
\end{equation} 
\aroundeqn
where each node stores an instance object $o \in O$, $g_i \in G_d$ denotes the corresponding dynamic Gaussians, and $m_o \in M$ is the transform matrix for each object. $ p_{o}(x_t, y_t, z_t) \in P$ is the center coordinate of the bounding box, and $a_o = (\theta_{t}, \phi_{t}) \in A $ is the orientation of the bounding box at time step $t \in T$.
Here, we compute Gaussians separately for each dynamic object. Using the transformation matrix $m_o$, we transform the coordinate system of the target object $o$ to the world coordinate where the static background resides:
\aroundeqn
\begin{equation}
  m_{o}^{-1} = R_{o}^{-1}S_{o}^{-1}
  \label{eq:object}
\end{equation}
\aroundeqn
where $R_{o}^{-1}$ and $S_{o}^{-1}$ are the rotation and translation matrices corresponding to each object. 

After optimizing all nodes in the dynamic Gaussian graph, we combine dynamic objects and static backgrounds using a Composite Gaussian Graph. Each node's Gaussian distribution is concatenated into the static Gaussian field based on the bounding box position and orientation in chronological order. In cases of occlusion between multiple dynamic objects, we adjust the opacity based on the distance from the camera center: closer objects have higher opacity, following the principles of light propagation:
\aroundeqn
\begin{equation}
  \alpha_{o, t} = \sum \frac{(p_{t}-b_{o})^{2} \cdot \cot a_o}{\| (b_o | R_o,S_o) - \rho \|^{2}} \alpha_{p_0}
  \label{eq:opacity}
\end{equation}
\aroundeqn
where $\alpha_{o, t}$ is the adjusted opacity of Gaussians for object $o$ at time step $t$, $p_{t} = (x_t, y_t, z_t)$ is the center of Gaussians for the object. $ [R_o, S_o] $ denotes the object-to-world transform matrix, $\rho$ denotes the center of camera view, and $\alpha_{p_{0}}$ is the opacity of Gaussians.

Finally, the composite Gaussian field, including both static background and multiple dynamic objects, can be formulated as:
\aroundeqn
\begin{equation}
  G_{comp} = \sum H <O, G_d, M, P, A, T> + G_{s}
  \label{eq:composite}
\end{equation}
\aroundeqn
where $G_s$ is obtained in Section~\ref{static} through Incremental Static 3D Gaussians and $H$ denots the optimized dynamic Gaussian graph.

\subsection{LiDAR Prior with surrounding views}
\label{LiDAR}
The primitive 3D-GS attempts to initialize Gaussians via structure-from-motion (SfM). However, unbounded urban scenes for autonomous driving contain many multiscale backgrounds and foregrounds. Nonetheless, they are only glimpsed through exceedingly sparse views, resulting in erroneous and incomplete recovery of geometric structures.

To provide better initialization for Gaussians, we introduce the LiDAR prior to 3D Gaussian to obtain better geometries and maintain multi-camera consistency in surrounding view registration. At each timestep $t \in T$, given a set of multi-camera images $ \{ I_{t}^{i} | i = 1 \ldots N \}$ collected from the moving platform and multi-frame LiDAR sweeps ${L_{t}}$. 
We aim to minimize multi-camera registration errors using LiDAR-image multi-modal data and obtain accurate point positions and geometric priors.

We first merge multiple frames of LiDAR sweeps to obtain the complete point cloud of the scene, denoted as $L$. We follow Colmap~\cite{schonberger2016structure} and extract image features $X = {x_p^q}$ from each image individually. Next, we project the LiDAR points onto surrounding images. For each LiDAR point $l$, we transform its coordinates to the camera coordinate system and match it with the 2D-pixel of the camera image plane through projection:
\aroundeqn
\begin{equation}
  x_p^q = K [R_t^{i} \cdot l_s + T_t^{i}]
  \label{eq:project}
\end{equation}
\aroundeqn
where $x_p^q$ is the 2D pixel of the image, $I_{t}^{i}$, $R_t^{i}$ and $T_t^{i}$ are orthogonal rotation matrices and translation vectors, respectively. $K \in \mathbb{R}^{3 \times 3}$ is the known camera intrinsic. Notably, points from LiDAR might be projected onto multiple pixels across multiple images. Therefore, we select the point with the shortest Euclidean distance to the image plane and retain it as the projected point, assigning color.

Similar to former works~\cite{schmied2023r3d3, herau2023moisst} in 3D reconstruction, we extend the dense bundle adjustment (DBA) to multi-camera setup and obtain the updated LiDAR points. Experiment results prove that initializing with LiDAR prior to aligning with surrounding multi-camera aids in providing the Gaussian model with more precise geometry priors.

 \begin{figure*}[ht]
  \centering
  \includegraphics[width=.9\linewidth]{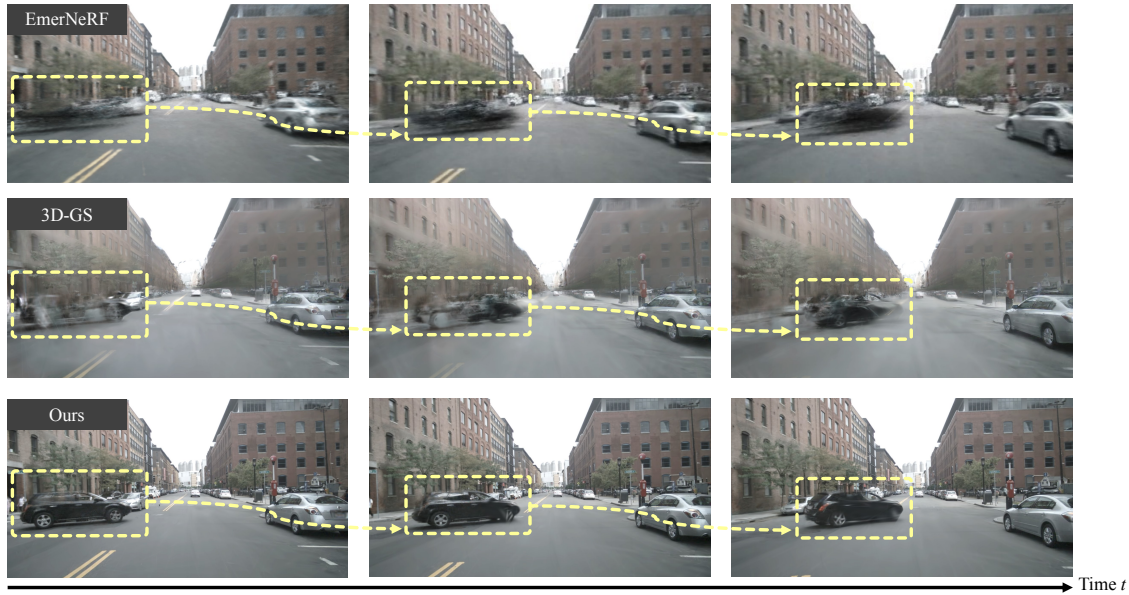}
  \caption{\textbf{Qualitative comparison on dynamic reconstruction.} We demonstrate the qualitative comparison results with our main competitors EmerNeRF~\cite{yang2023emernerf} and 3D-GS~\cite{kerbl20233d} on dynamic reconstruction for 4D driving scenes of nuScenes. \ourmethod{} enables the high-quality reconstruction of dynamic objects at high speed while maintaining temporal consistency.}
  \label{temporal_compare}
  \afterfig
 \end{figure*}

\subsection{Global Rendering via Gaussian Splatting}
\label{rendering}
We adopt the differentiable 3D Gaussian splatting renderer $\varsigma$ from~\cite{kerbl20233d} and project the global composite 3D Gaussian into 2D, where the covariance matrix $\widetilde{\Sigma}$ is given by:
\aroundeqn
\begin{equation}
  \widetilde{\Sigma} = J E \ \Sigma \ E^\top J^\top
  \label{eq:cocariance}
\end{equation}
\aroundeqn
where $J$ is the Jacobian matrix of the perspective projection, and $E$ denotes the world-to-camera matrix.

The composite Gaussian field projects the global 3D Gaussian onto multiple 2D planes and is supervised using surrounding views at each time step. In the global rendering process, Gaussians from the next time step are initially invisible to the current and subsequently incorporated with the supervision of corresponding global images.

The loss function of our method consists of three parts. Following~\cite{kerbl20233d, xie2023s3im}, we first introduce the Tile Structural Similarity (TSSIM) to Gaussian Splatting, which measures the similarity between the rendered tile and the corresponding ground truth. 
\aroundeqn
\begin{equation}
  L_{TSSIM}(\delta) = 1 - \frac{1}{Z}\sum_{z=1}^{Z} SSIM(\Psi(\hat{C}), \Psi(C))
  \label{eq:TSSIM loss}
\end{equation}
\aroundeqn
where we split the screen into $M$ tiles, $\delta$ is the training parameters of the Gaussians, $\Psi(\hat{C})$ denotes the rendered tile from Composite Gaussian Splatting, and $\Psi(C)$ denotes the paired ground-truth tile.

We also introduce robust loss for reducing outliers in 3D Gaussians, which can be defined as:
\aroundeqn
\begin{equation}
  L_{Robust}(\delta) = \kappa(\|\hat{I} - I\|_{2})
  \label{eq:robust loss}
\end{equation}
\aroundeqn
where $\kappa \in (0,1]$ is the shape parameter that controls the robustness of the loss, $I$ and $\hat{I}$ denote the ground truth and synthesis image, respectively.

The LiDAR loss is further employed by supervising the expected Gaussians' position from the LiDAR, obtaining better geometric structure and edge shapes:
\aroundeqn
\begin{equation}
  L_{LiDAR}(\delta) = \frac{1}{s}\sum \|P(G_{comp}) - L_s\|^{2}
  \label{eq:LiDAR loss}
\end{equation}
\aroundeqn
where $P(G_{comp})$ is the position of 3D Gaussians, and $L_s$ is the LiDAR point prior. We optimize the Composite Gaussians by minimizing the sum of three losses.
\section{Experiments}
\label{sec:expe}

\begin{table*}[!t]
  \caption{
  {\textbf{Overall perforamnce of \ourmethod{} with existing state-of-the-art approaches on the nuScenes dataset.} Ours-S denotes the \ourmethod{} with SfM initialization, and Ours-L denotes training the Gaussian model with LiDAR prior.}}
  \centering
  \footnotesize
  \setlength{\tabcolsep}{7mm}{
    \begin{tabular}{ccccc}
 \toprule
    \textbf{Methods} & \textbf{Input} & \textbf{PSNR $\uparrow$} & \textbf{SSIM $\uparrow$} & \textbf{LPIPS $\downarrow$} \\
    \hline 
    Instant-NGP~\cite{muller2022instant}           & Images         & 16.78 & 0.519 & 0.570  \\
    NeRF+Time                              & Images         & 17.54 & 0.565 & 0.532  \\
    Mip-NeRF~\cite{barron2021mip}          & Images         & 18.08 & 0.572 & 0.551  \\
    Mip-NeRF360~\cite{barron2022mip}       & Images         & 22.61 & 0.688 & 0.395  \\
    Urban-NeRF~\cite{rematas2022urban}     & Images + LiDAR & 20.75 & 0.627 & 0.480  \\
    S-NeRF~\cite{xie2023s}                 & Images + LiDAR & 25.43 & 0.730 & 0.302  \\
    SUDS~\cite{turki2023suds}              & Images + LiDAR & 21.26 & 0.603 & 0.466  \\
    EmerNeRF~\cite{yang2023emernerf}       & Images + LiDAR & \cellcolor{tabthird} 26.75 & \cellcolor{tabthird} 0.760 & 0.311  \\
    \hline
    3D-GS~\cite{kerbl20233d}               & Images + SfM Points & 26.08 & 0.717 & \cellcolor{tabthird} 0.298  \\
    \hline
    Ours-S                                 & Images + SfM Points & \cellcolor{tabsecond} 28.36 & \cellcolor{tabsecond} 0.851 & \cellcolor{tabsecond} 0.256  \\
    \textbf{Ours-L}                        & Images + LiDAR      & \cellcolor{tabfirst} 28.74  & \cellcolor{tabfirst} 0.865 & \cellcolor{tabfirst} 0.237  \\
    \hline
    \end{tabular}%
    }
  \label{compare_SOTA}%
  \aftertab
\end{table*}%

\subsection{Datasets}

The nuScenes~\cite{caesar2020nuscenes} dataset is a public large-scale dataset for autonomous driving, containing 1000 driving scenes collected with multiple sensors (6 cameras, 1 LiDAR, etc). It has annotations of 23 object classes with accurate 3D bounding boxes. Our experiment uses the keyframes of 6 challenging scenes with surrounding views collected from 6 cameras and corresponding LiDAR sweeps (optional) as input.
The KITTI-360~\cite{liao2022kitti} dataset contains multiple sensors, corresponding to over 320k images and point clouds. Even though the dataset provides stereo camera images, we only use a single camera to demonstrate that our method also performs well in monocular scenes.

\subsection{Implementation Details}
Our implementation is primarily based on the 3D-GS framework, with fine-tuned optimization parameters to fit the large-scale unbounded scenes. Instead of using SfM points or randomly initialized points as input, we employ the LiDAR prior mentioned in Section~\ref{LiDAR} as initializations. Considering the computational cost, we use a voxel grid filter for LiDAR points, reducing the scale without losing geometric features. We employ random initialization for dynamic objects with initial points set to 3000, given that objects are relatively small in large-scale scenes. We increase the total training iterations to 50,000, set the threshold for densifying grad to 0.001, and reset the opacity interval to 900. The learning rate of Incremental Static 3D Gaussians remains the same as in the official setting, while the learning rate of the Composite Dynamic Gaussian Graph exponentially decays from 1.6e-3 to 1.6e-6. All the experiments are carried out on 8 RTX8000 with 384 GB memory in total.

\begin{table}[!t]
  \caption{
  {\textbf{Overall perforamcne of \ourmethod{} with existing state-of-the-art approaches on the KITTI-360 dataset.}}}
  \centering
  \footnotesize
  \setlength{\tabcolsep}{5.5mm}{
    \begin{tabular}{ccc}
    \toprule
    \textbf{Methods} & \textbf{PSNR $\uparrow$} & \textbf{SSIM $\uparrow$} \\
    \hline 
    NeRF~\cite{mildenhall2021nerf}        & 21.94 & 0.781  \\
    Point-NeRF~\cite{xu2022point}         & 21.54 & 0.793   \\
    NSG~\cite{ost2021neural}              & 22.89 & 0.836   \\
    Mip-NeRF360~\cite{barron2022mip}      & 23.27 & 0.836   \\
    SUDS~\cite{turki2023suds}             & 23.30 & 0.844   \\
    DNMP~\cite{lu2023urban}               & \cellcolor{tabthird} 23.41 & \cellcolor{tabthird} 0.846   \\
    \hline
    Ours-S                                & \cellcolor{tabsecond} 25.18 & \cellcolor{tabsecond} 0.862   \\
    \textbf{Ours-L}                       & \cellcolor{tabfirst} 25.62  & \cellcolor{tabfirst} 0.868  \\
    \hline
    \end{tabular}%
    }
  \label{KITTI-360}%
  \aftertab
\end{table}%

\subsection{Results and Comparisons}

\noindent \textbf{Comparisons of surrounding views synthesis on nuScenes.}
We conduct benchmarking against the state-of-the-art approaches, including NeRF-based methods~\cite{muller2022instant, barron2021mip, barron2022mip, rematas2022urban, xie2023s, turki2023suds, yang2023emernerf} and 3DGS-based method~\cite{kerbl20233d}.

As shown in Table~\ref{compare_SOTA}, our method outperforms Instant-NGP~\cite{muller2022instant} by a large margin, which employs a hash-based NeRF for novel view synthesis. Mip-NeRF~\cite{barron2021mip} and Mip-NeRF360~\cite{barron2022mip} are two methods designed for unbounded outdoor scenes. Our method also significantly surpasses them across all evaluated metrics. 

Urban-NeRF~\cite{rematas2022urban} first introduces LiDAR to NeRF to reconstruct urban scenes. However, it primarily only leverages LiDAR to provide depth supervision. Instead, we leverage LiDAR as a more accurate geometric prior and incorporate it into Gaussian models, proven more effective for large-scale scene reconstruction. Our proposed method achieves superior results compared to S-NeRF~\cite{xie2023s} and SUDS~\cite{turki2023suds}, both of which decompose the scene into static background and dynamic objects and construct the scene with the help of LiDAR.
Compared to our primary competitor, EmerNeRF~\cite{yang2023emernerf}, which applies spatial-temporal representation for dynamic driving scenes using flow fields. Our method outperforms it across all metrics, eliminating the necessity for estimating scene flow.
For Gaussian-based approaches, our method boosts the performance of our baseline method 3D-GS~\cite{kerbl20233d} on large-scale scenes across all evaluated metrics and achieves optimal results.

We also compare qualitatively with our main competitors EmerNeRF~\cite{yang2023emernerf} and 3D-GS~\cite{kerbl20233d} on challenging nuScenes driving scenes. For surrounding view synthesis with multi-camera, as shown in Figure~\ref{camera_compare}, our method enables the generation of photo-realistic rendering images and ensures view consistency across multi-camera. Meanwhile, EmerNeRF~\cite{yang2023emernerf} and 3D-GS~\cite{kerbl20233d} struggle in challenging regions, displaying undesirable visual artifacts such as ghosting, dynamic object disappearance, loss of plant texture details, lane markings, and distant scene blurring.  

We further demonstrate the reconstruction results for dynamic temporal scenes.
Our method accurately models dynamic objects within large-scale scenes, mitigating issues such as loss, ghosting, or blurring of these dynamic elements. We also maintain consistency in constructing dynamic objects over time, even when they move at a relatively fast speed. In comparison, ~\cite{yang2023emernerf, kerbl20233d} both fail in modeling fast-moving dynamic objects, as shown in Figure~\ref{temporal_compare}.

\noindent \textbf{Comparisons of mono-view synthesis on KITTI-360.}
To further validate the effectiveness of our method on monocular driving scene setting, we conduct experiments with the KITTI-360 dataset and compare it with existing SOTA methods, including NeRF-based methods NeRF~\cite{mildenhall2021nerf}, Mip-NeRF360~\cite{barron2022mip}, point-based method Point-NeRF~\cite{xu2022point}, graph-based method NSG~\cite{ost2021neural}, flow-based method SUDS~\cite{turki2023suds}, and mesh-based method DNMP~\cite{lu2023urban}. As shown in Table~\ref{KITTI-360}, our method demonstrates the optimal performance in monocular driving scenes, surpassing existing methods by a large margin.  More results and videos are available in the supplementary materials.

\subsection{Ablation Study}

\textbf{Initialization prior for Gaussians.} Comparative experiments are conducted to analyze the effect of different priors and initialization methods on the Gaussian model. The original 3D-GS provides two initialization modes: randomly generated points and SfM points computed by COLMAP~\cite{schonberger2016structure}. We additionally offer two other initialization approaches: point clouds exported from a pre-trained NeRF model and points generated with LiDAR prior. 

Meanwhile, to analyze the effect of point cloud quantity, we down-sample the LiDAR to 600K and apply adaptive filtering (1M) to control the number of generated LiDAR points. We also set different maximum thresholds for randomly generated points (600K and 1M). Here, SfM-600K$\pm$20K represents the points number computed by COLMAP, NeRF-1M$\pm$20K denotes the total points generated by the pre-trained NeRF model, and LiDAR-2M$\pm$20k refers to the original quantity of LiDAR points.

\begin{figure}[!t]
  \centering
  \includegraphics[width=0.7\linewidth]{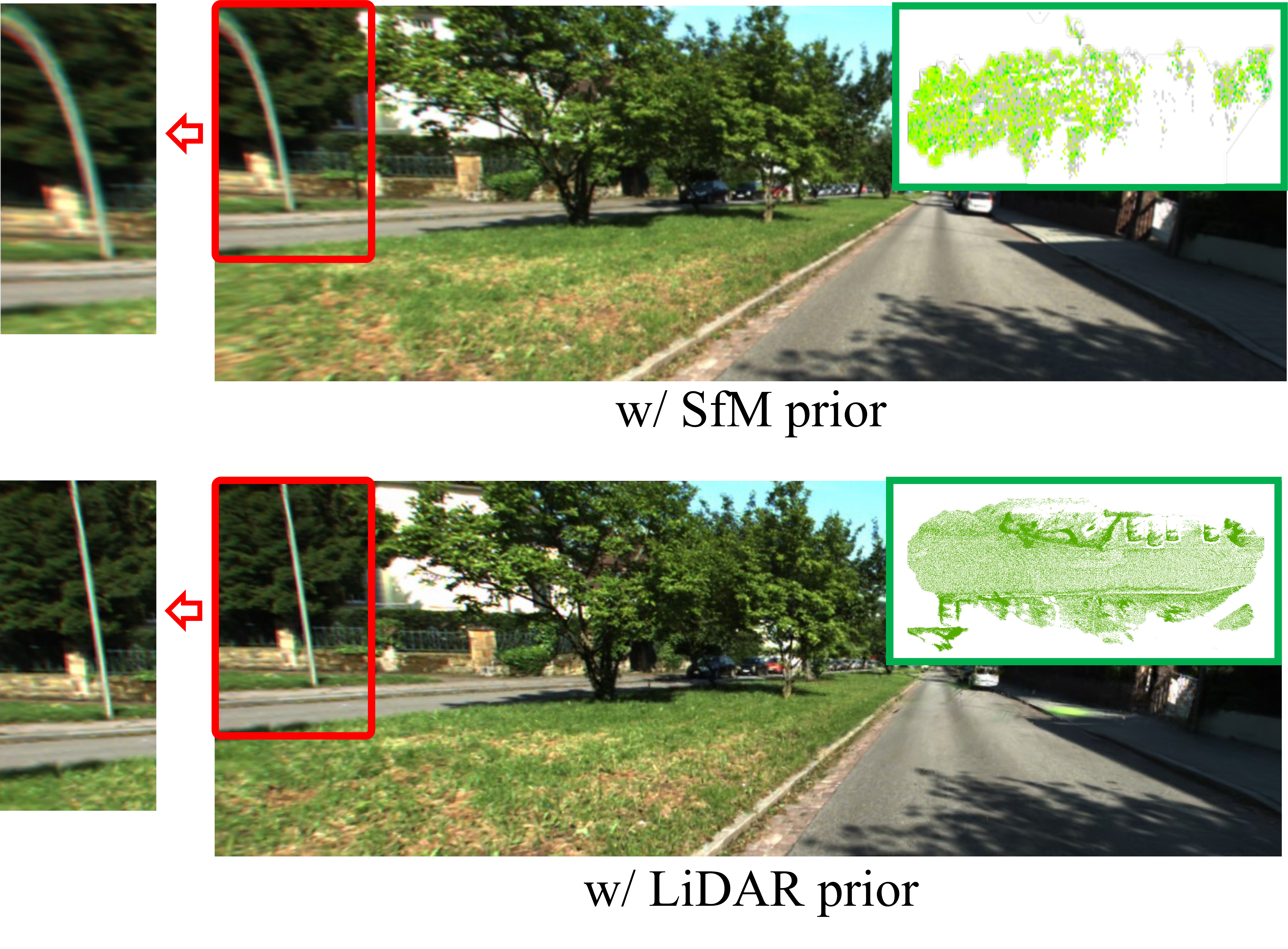}
  \caption{\textbf{Visualization comparison using different initialization methods on KITTI-360.} Compared to initialization with SfM points~\cite{kerbl20233d}, using LiDAR prior allows Gaussians to restore more accurate geometric structures in the scene.}
  \label{fig:effect-LiDAR}
  \afterfig
 \end{figure}

As shown in Table~\ref{LiDAR_prior}, randomly generated points lead to the worst results as they lack any geometric prior. Initializing with SfM points also cannot adequately recover the scene's precise geometries due to the sparse points and intolerable structural errors. Leveraging point clouds generated from a pre-trained NeRF model provides a relatively accurate geometric prior, but there are still noticeable outliers. For the model initialized with LiDAR prior, although downsampling results in loss of geometric information in some local regions, it still retains relatively accurate structural priors, thus surpassing SfM (Figure~\ref{fig:effect-LiDAR}). We can also observe that experiment results do not linearly change with increasing LiDAR point quantities. We deduce this is because overly dense points store redundant features that interfere with the optimization of the Gaussian model.

\begin{table}[!t]
  \centering
  \caption{
  {\textbf{Effect of different initialization methods on the Gaussian model.} LiDAR-600K $\dagger$ denotes for downsampling the original LiDAR data to a corresponding point cloud magnitude. LiDAR-1M  $\ddagger$ denotes denoising and removing outliers in LiDAR points, which is used in our method.}}
    \footnotesize
  \centering
  \setlength{\tabcolsep}{1.4mm}{
  {
    \begin{tabular}{cccc}
   \toprule
    \textbf{Methods} & \textbf{PSNR $\uparrow$} & \textbf{SSIM $\uparrow$} & \textbf{LPIPS $\downarrow$} \\
    \hline 
    Random-600K   & 22.18     & 0.653     & 0.424  \\
    Random-1M     & 22.23     & 0.653     & 0.421  \\
    SfM-600K      & 28.36     & 0.851     & 0.256  \\
    NeRF-1M       & \cellcolor{tabthird} 28.51     & \cellcolor{tabthird} 0.858     & \cellcolor{tabthird} 0.251  \\
    \hline
    LiDAR-600K $\dagger$      & 28.49     & 0.854     & \cellcolor{tabsecond} 0.245 \\
    LiDAR-1M  $\ddagger$      & \cellcolor{tabsecond} 28.74     & \cellcolor{tabsecond} 0.865     & \cellcolor{tabfirst} 0.237  \\
    LiDAR-2M                  & \cellcolor{tabfirst} 28.78     & \cellcolor{tabfirst} 0.867     & \cellcolor{tabfirst} 0.237  \\
    \hline
    \end{tabular}%
    }
    }
  \label{LiDAR_prior}%
  \aftertab
\end{table}%

\noindent \textbf{Effectiveness of Each Module.}
We analyze how each proposed module contributes to the final performance. As shown in Table~\ref{Ablation_Study}, the Composite Dynamic Gaussian Graph module plays a crucial role in reconstructing dynamic driving scenes, while the Incremental Static 3D Gaussians module enables high-quality large-scale background reconstruction. These two novel modules significantly enhance the modeling quality of complex driving scenes. Regarding the proposed loss functions, results indicate that both $L_{TSSIM}$ and $L_{Robust}$ notably improve the rendering quality, enhancing texture details and removing artifacts. $L_{LiDAR}$, assisted by LiDAR prior, helps Gaussians achieve better geometric priors. Experimental results also demonstrate that \ourmethod{} performs well even without LiDAR prior, showcasing strong robustness for various initialization methods.

\begin{table}[!t]
  \centering
  \caption{
  {\textbf{Effect of each module in our proposed method.} IS3G is short for the Incremental Static 3D Gaussians module, and CDGG is short for the Composite Dynamic Gaussian Graph module.}}
    \footnotesize
  \centering
  \setlength{\tabcolsep}{1.4mm}{
  {
    \begin{tabular}{cccc}
 \toprule
    \textbf{Model} & \textbf{PSNR $\uparrow$} & \textbf{SSIM $\uparrow$} & \textbf{LPIPS $\downarrow$} \\
    \hline 
    w/o IS3G             & 27.72     & 0.771     & 0.295  \\
    w/o CDGG             & 26.97     & 0.752     & 0.306  \\
    w/o $L_{TSSIM}$      & 27.88     & 0.783     & 0.280  \\
    w/o $L_{Robust}$     &  28.05     &  0.814     & 0.271  \\
    w/o $L_{LiDAR}$      & \cellcolor{tabsecond} 28.45     & \cellcolor{tabsecond} 0.854     & \cellcolor{tabsecond} 0.248  \\
    Ours-S               & \cellcolor{tabthird} 28.36 & \cellcolor{tabthird} 0.851 & \cellcolor{tabthird} 0.256  \\
    Ours-L               & \cellcolor{tabfirst} 28.74  & \cellcolor{tabfirst} 0.865 & \cellcolor{tabfirst} 0.237  \\
    \hline
    \end{tabular}%
    }
    }
  \label{Ablation_Study}%
  \aftertab
\end{table}%

\subsection{Corner Case Simulation}
\begin{figure}[h]
  \centering
  \includegraphics[width=.9\linewidth]{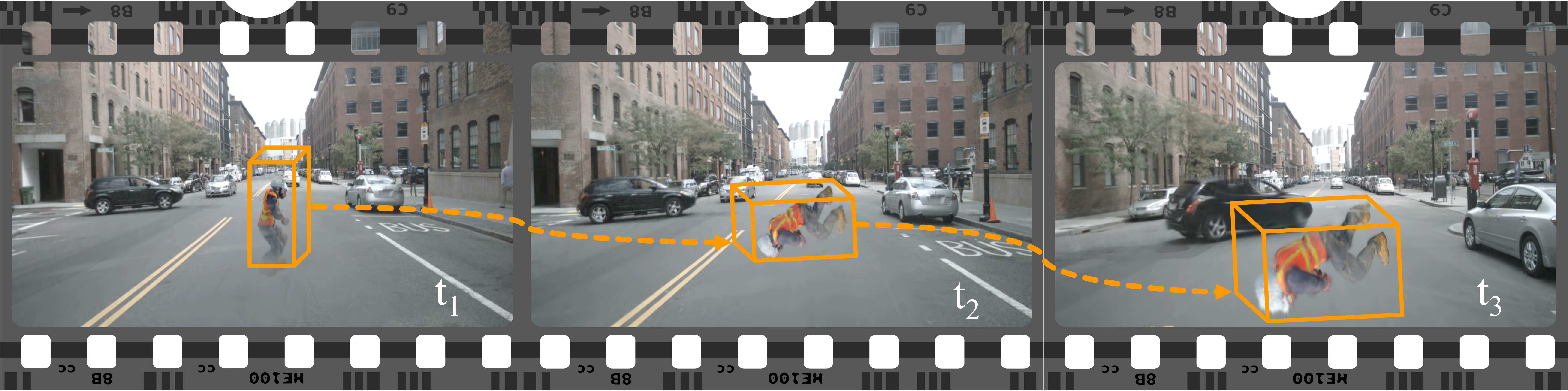}
  \caption{\textbf{Example of corner case simulation.} Corner case simulation using DrivingGaussian: A man walking on the road suddenly falls, and a car approaches ahead.}
  \label{fig:editing}
  \vspace{-10pt}
\end{figure}

We demonstrate the effectiveness of our approach to simulating corner cases in real-world driving scenes. As shown in Figure~\ref{fig:editing}, we can insert arbitrary dynamic objects into the reconstructed Gaussian field. The simulated scene maintains temporal coherence and exhibits good inter-sensor consistency among multiple sensors. Our method enables controllable simulation and editing for autonomous driving scenes, facilitating safe self-driving systems research.

\section{Conclusion}
\label{sec:con}

We introduce \ourmethod{}, a novel framework for representing large-scale dynamic autonomous driving scenes based on the proposed Composite Gaussian Splatting.
\ourmethod{} progressively models the static background with incremental static 3D Gaussians and captures multiple moving objects using a composite dynamic Gaussian graph.
We further leverage LiDAR prior for accurate geometric structures and multi-view consistency.
\ourmethod{} achieves state-of-the-art performance on two autonomous driving datasets, allowing high-quality surrounding view synthesis and dynamic scene reconstruction.
\section{Acknowledgment}
\label{ack}
This work was supported by National Key R\&D Program of China (Grant No.2022ZD0160305). This work was also a research outcome of Key Laboratory of Science, Technology and Standard in Press Industry (Key Laboratory of Intelligent Press Media Technology).

{
    \small
    \bibliographystyle{ieeenat_fullname}
    \bibliography{main}
}


\clearpage
\setcounter{page}{1}
\maketitlesupplementary

\section{Implementation Details}
\label{details}

\noindent \textbf{Experimental Details.}
We assess our models using various metrics, including PSNR, SSIM, and LPIPS. We report the average results of all camera frames on the selected scenes. For the nuScenes~\cite{caesar2020nuscenes} dataset, images of full-resolution $1600 \times 900$ are rendered with 360-degree horizontal FOV per time-step. We use synchronized images from 6 cameras in surrounding views as inputs. We randomly select every 5th image of different cameras in the sequences as the test set and utilize the remaining images as the training set. For the KITTI-360~\cite{liao2022kitti} dataset, we only use sequential images from a single camera as input, with a resolution of $1408 \times 376$. We select every 10th image of the camera in the sequences as the test set.

\begin{figure}[!t]
  \centering
  \includegraphics[width=\linewidth]{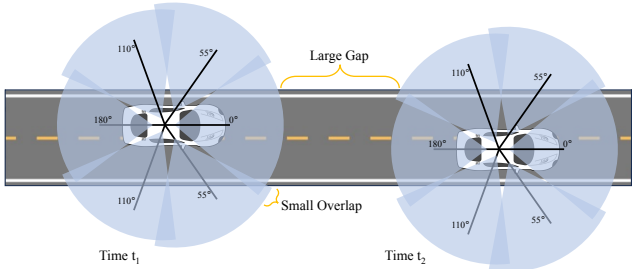}
  \caption{\textbf{Visualization of surrounding multi-camera views in nuScenes dataset.} The surrounding views have small overlaps among multi-camera but large gaps across time.}
  \label{multi-camera}
  \afterfig
 \end{figure}

\noindent \textbf{Details of LiDAR Prior.}
The LiDAR prior provides more precise and complete initialization oversight for scene modeling, helping to recover the more correct and detailed shape of the scene. Here, we present detailed preprocessing and techniques for using the LiDAR prior. 

LiDAR points derived from the dataset are categorized into dynamic foreground and static background. Dynamic foreground can cause misalignment during LiDAR-image registration due to drag, aliasing, etc. So, we first cut out dynamic objects from the LiDAR points based on the segmentation labels, obtaining purely static LiDAR prior to the scenes. We then use multi-frame aggregation to stitch together the LiDAR of the scene according to the currently visible regions of the Incremental Static 3D Gaussians. The coordinates of LiDAR prior are further transformed into the global coordinate system via calibration matrices.

Intuitively, while capturing images with moving platforms, nearby areas will have more pixels to represent finer details. In contrast, distant regions are described using a limited number of coarse points. This principle similarly applies to the 3D Gaussian representation for large-scale driving scenes. In this regard, we utilize an adaptive filtering algorithm to optimize the LiDAR prior. The previously obtained LiDAR point cloud is voxelized into a fixed-size voxel grid. We divide the voxel grid along the rays extending forward from the camera center based on depth. We next apply distance weighting and remove isolated outliers for the points within the voxel grids representing distant views.

\noindent \textbf{Details of surrounding multi-camera views.}
As shown in Figure~\ref{multi-camera}, we show the distribution of the surrounding multi-camera views in driving scenes~\cite{caesar2020nuscenes}. We can observe that these surrounding views have only minimal overlap between multi-camera but have large intervals across adjacent frames. Compared with typical NeRF-based captures (e.g., central objects captured by hemisphere views), the surrounding multi-camera views pose a great challenge to modeling the whole scene from such sparse observations.  

\begin{figure}[!t]
  \centering
  \includegraphics[width=\linewidth]{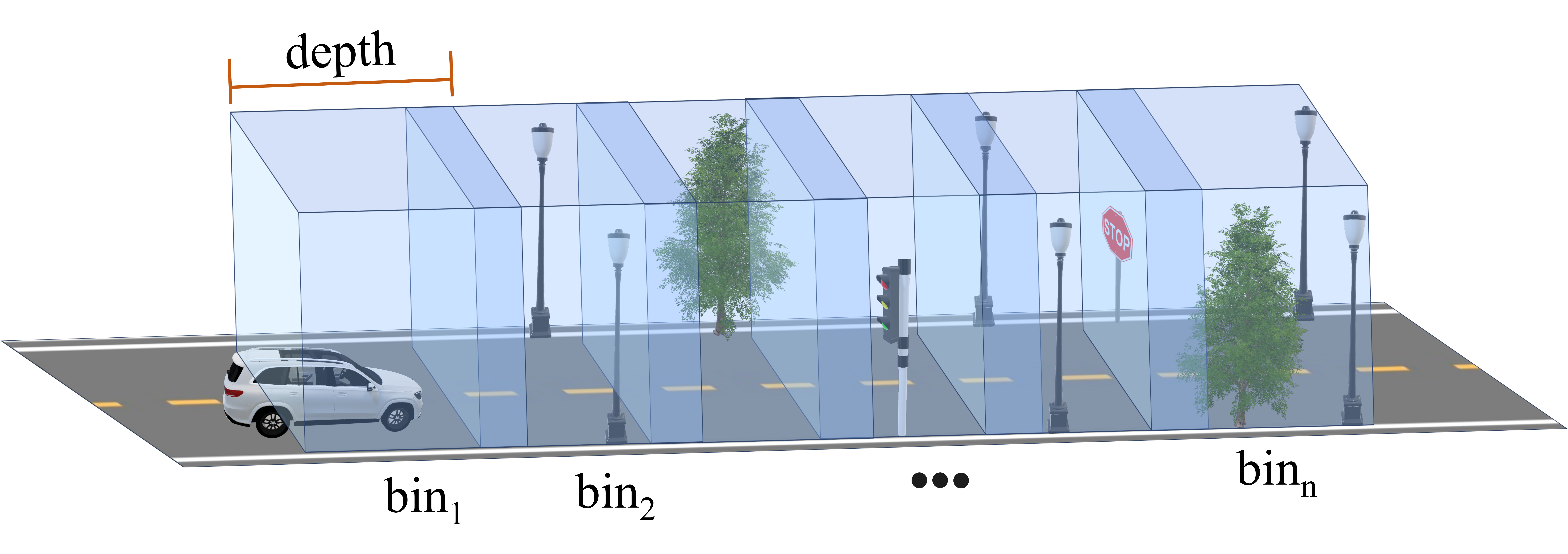}
  \caption{\textbf{Visualization of bins arrangement for the Incremental Static 3D Gaussians.} The small overlap between two neighboring bins is used to align the static backgrounds of the two bins.}
  \label{bins}
  \afterfig
 \end{figure}

\noindent \textbf{Details of bins arrangement for static background.}
We show the arrangement of sequential bins in the Incremental Static 3D Gaussians module. As shown in Figure~\ref{bins}, each bin is distributed according to the scene's depth and contains one or more frames of surrounding images. Neighboring bins have a small overlap region, which is used to align the static backgrounds of two bins. The latter bin is then incrementally fused into the Gaussian field of the previous bins. In addition, we allow the distribution of bins to be specified manually, enabling better adaptation to extreme or depth-unknown scenarios.

\begin{figure}[ht]
  \centering
  \includegraphics[width=\linewidth]{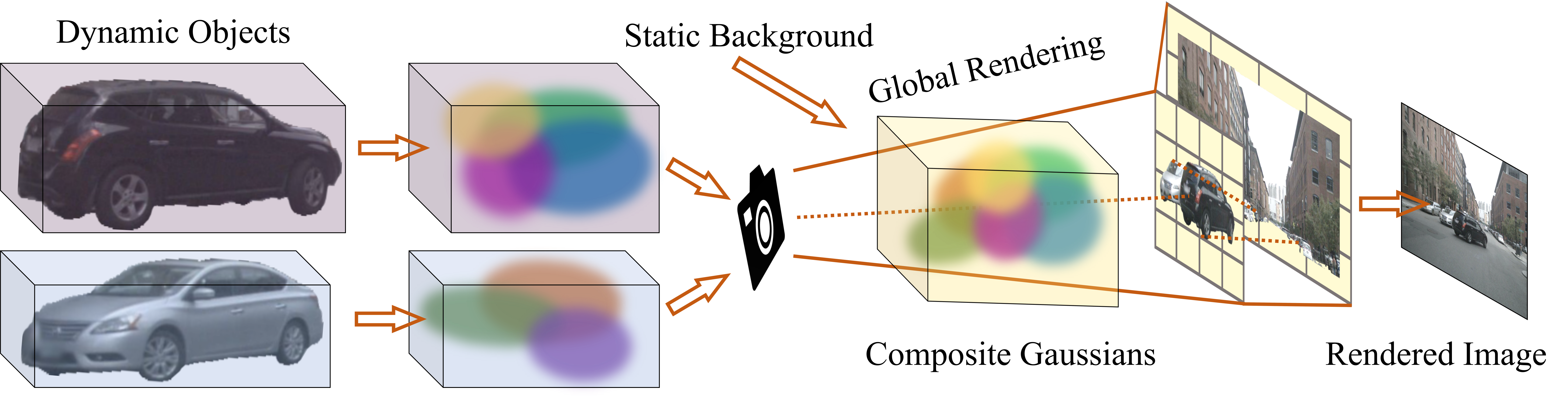}
  \caption{\textbf{Visualization of Global Rendering vis GS.} \ourmethod{} ensures the reconstruction of multiple dynamic objects along with their accurate positions and occlusion relationships.}
  \label{global}
  \vspace{-8pt}
  \afterfig
 \end{figure}

\begin{table*}[!t]
  \caption{
  {\textbf{Overall perforamnce of \ourmethod{} with existing state-of-the-art approaches on the nuScenes dataset.} Ours-S denotes the \ourmethod{} with SfM initialization, and Ours-L denotes training the Gaussian model with LiDAR prior. Rendering Time denotes the rendering time for each frame.}}
  \centering
  \footnotesize
  \setlength{\tabcolsep}{5.5mm}{
    \begin{tabular}{cccccc}
 \toprule
    \textbf{Methods} & \textbf{Input} & \textbf{PSNR $\uparrow$} & \textbf{SSIM $\uparrow$} & \textbf{LPIPS $\downarrow$} & \textbf{Rendering Time(s)}\\
    \hline 
    Instant-NGP~\cite{muller2022instant}           & Images         & 16.78 & 0.519 & 0.570 & 4.382 \\
    NeRF+Time                              & Images         & 17.54 & 0.565 & 0.532 & 31.14 \\
    Mip-NeRF~\cite{barron2021mip}          & Images         & 18.08 & 0.572 & 0.551 & 24.55 \\
    Mip-NeRF360~\cite{barron2022mip}       & Images         & 22.61 & 0.688 & 0.395 & 11.86 \\
    NSG~\cite{ost2021neural}               & Images         & 21.67     & 0.671     & 0.424 &  52.28 \\
    Urban-NeRF~\cite{rematas2022urban}     & Images + LiDAR & 20.75 & 0.627 & 0.480 & 41.29 \\
    S-NeRF~\cite{xie2023s}                 & Images + LiDAR & 25.43 & 0.730 & 0.302 & 23.67 \\
    SUDS~\cite{turki2023suds}              & Images + LiDAR & 21.26 & 0.603 & 0.466  & 45.7 \\
    EmerNeRF~\cite{yang2023emernerf}       & Images + LiDAR & \cellcolor{tabthird} 26.75 & \cellcolor{tabthird} 0.760 & 0.311 & 21.91 \\
    \hline
    3D-GS~\cite{kerbl20233d}               & Images + SfM Points & 26.08 & 0.717 & \cellcolor{tabthird} 0.298 & \cellcolor{tabfirst} 0.864 \\
    4D-GS~\cite{wu20234d}                  & Images + SfM Points & 19.79 & 0.622 & 0.473 & 2.160  \\
    \hline
    Ours-S                                 & Images + SfM Points & \cellcolor{tabsecond} 28.36 & \cellcolor{tabsecond} 0.851 & \cellcolor{tabsecond} 0.256 & \cellcolor{tabthird} 0.965 \\
    \textbf{Ours-L}                        & Images + LiDAR      & \cellcolor{tabfirst} 28.74  & \cellcolor{tabfirst} 0.865 & \cellcolor{tabfirst} 0.237 & \cellcolor{tabsecond} 0.963 \\
    \hline
    \end{tabular}%
    }
  \label{compare_SOTA2}%
  \aftertab
\end{table*}%

\noindent \textbf{Moving Objects in Driving Scenes.}
Dynamic objects are foreground instances that are moving in the current scene, while parked vehicles or static objects are not. We provide two methods of decoupling dynamic objects for our approach, using either a 3D bounding box or pre-trained object segmentation foundation models (e.g., Grounded SAM~\cite{ren2024grounded}, SEEM~\cite{zou2023segment} or OmniMotion~\cite{wang2023tracking}). 

Using the 3D bounding box, we project the bounding box of each object individually onto 2D images of the surrounding view and mask the objects inside the box. We explicitly align the dynamic objects in each frame with the ID of each object from the label.

Similarly, when using pre-trained dynamic object segmentation models, we separate dynamic objects from static areas by applying pre-trained models and explicitly labeling each object individually with the object ID. Experiments also show that it is unnecessary to precisely segment every pixel while excluding background pixels, as our method is robust to dynamic objects containing background pixels. These imperfect background pixels are eliminated in the optimization of modeling dynamic objects in the scene.

\noindent \textbf{Details of Global Rendering vis GS.}
Global rendering vis GS aims to restore the position relationship and occlusion of multiple dynamic objects with static backgrounds in the real driving scene. We utilize the fast splatting algorithm introduced by the 3D-GS~\cite{kerbl20233d} to support the global rendering. As shown in Figure~\ref{global}, our method enables the re-rendering of multiple objects and static backgrounds in a shared driving scene. Based on the explicit geometry scene structure of Gaussian distribution, the global rendering preserves the original occlusion relationships and exact spatial positions.

\begin{figure*}[ht]
  \centering
  \includegraphics[width=\linewidth]{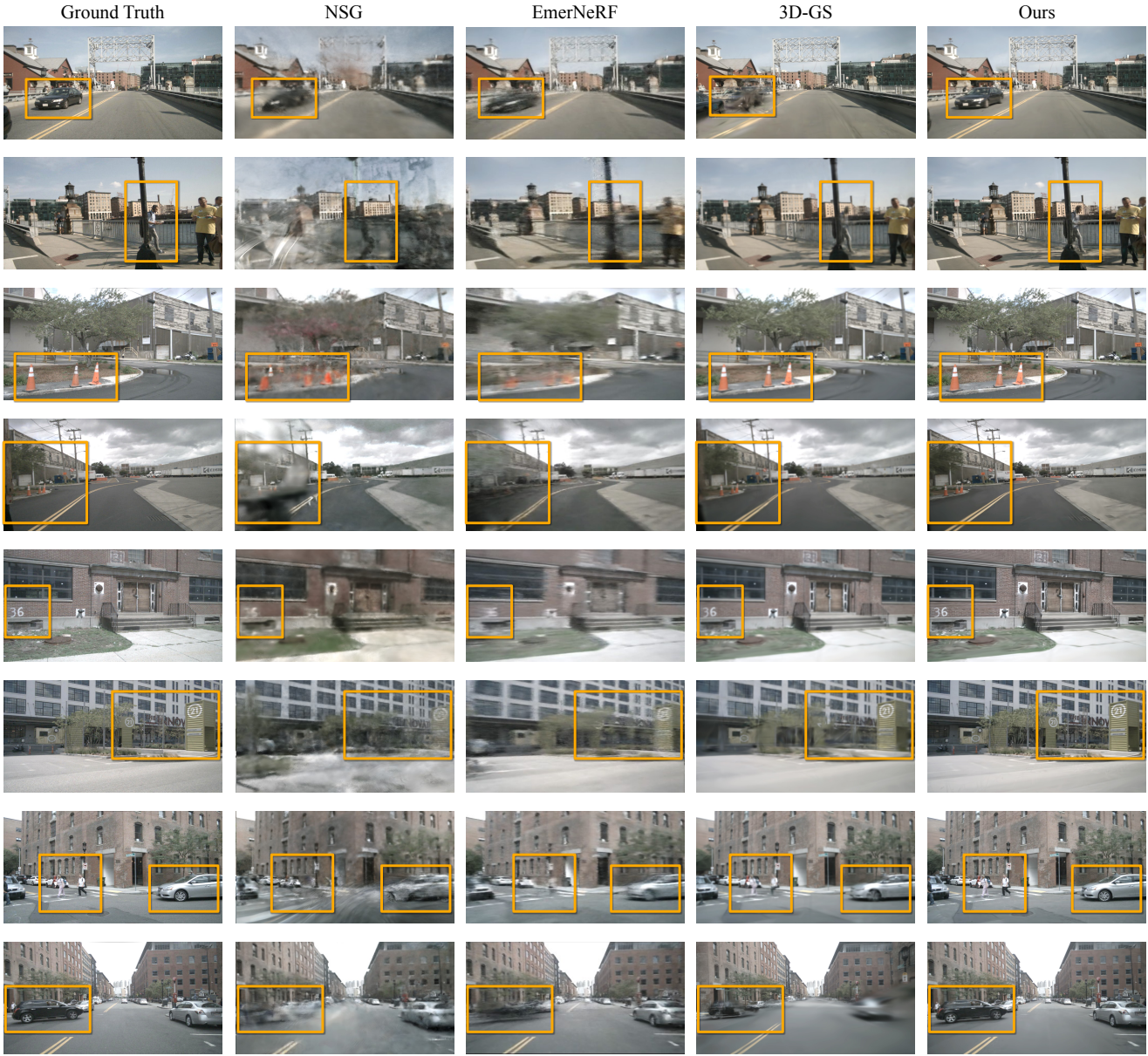}
  \caption{\textbf{Qualitative comparison on the nuScenes dataset.} We demonstrate the qualitative comparison results with our main competitors NSG~\cite{ost2021neural}, EmerNeRF~\cite{yang2023emernerf} and 3D-GS~\cite{kerbl20233d} on driving scenes reconstruction of nuScenes.}
  \label{nuScenes_comp}
  \afterfig
 \end{figure*}

\section{Additional Results on nuScenes}
\label{nuscenes}

\noindent \textbf{Quantitative Comparison.}
We provide more comparisons of results with recent works on large-scale driving scenes and 3D Gaussian-based approaches. For a fair comparison, we also migrate the graph-based method NSG~\cite{ost2021neural} and dynamic Gaussians method 4D-GS~\cite{wu20234d} to the nuScenes dataset. As shown in Table~\ref{compare_SOTA2}, our method boosts the performance of NSG across three metrics. Although NSG similarly uses a graph-based representation for dynamic objects, it only applies to the front-forward monocular views and does not cope well with the dynamic objects under ego vehicle movements. Our method also shows a huge lead compared to the latest work designed for dynamic 3D Gaussians~\cite{wu20234d}. Since 4D-GS~\cite{wu20234d} employs Gaussians updated over time steps to represent dynamic objects, it only works for slow-moving central objects and fails in complex scenes with multiple high-speed moving foregrounds.

\noindent \textbf{Rendering Speed.}
Table~\ref{compare_SOTA2} shows that our method achieves a good balance between rendering quality and rendering speed. Compared to the accelerated NeRF method Instant-NGP~\cite{muller2022instant}, our approach achieves higher results with faster rendering speed. Our method achieves the optimal quality with less rendering time compared to those of NeRF-based methods designed for unbounded large-scale scenes(e.g., Mip-NeRF~\cite{barron2021mip}, Mip-NeRF360~\cite{barron2022mip}, Urban-NeRF~\cite{rematas2022urban}). Compared to methods~\cite{xie2023s, turki2023suds, yang2023emernerf}, also designed for dynamic driving scenarios, our approach undoubtedly obtains the best performance and a significant reduction in rendering time. Compared to our baseline method 3D-GS~\cite{kerbl20233d}, our method achieves higher rendering quality with comparable rendering speeds.

\begin{figure*}[ht]
  \centering
  \includegraphics[width=\linewidth]{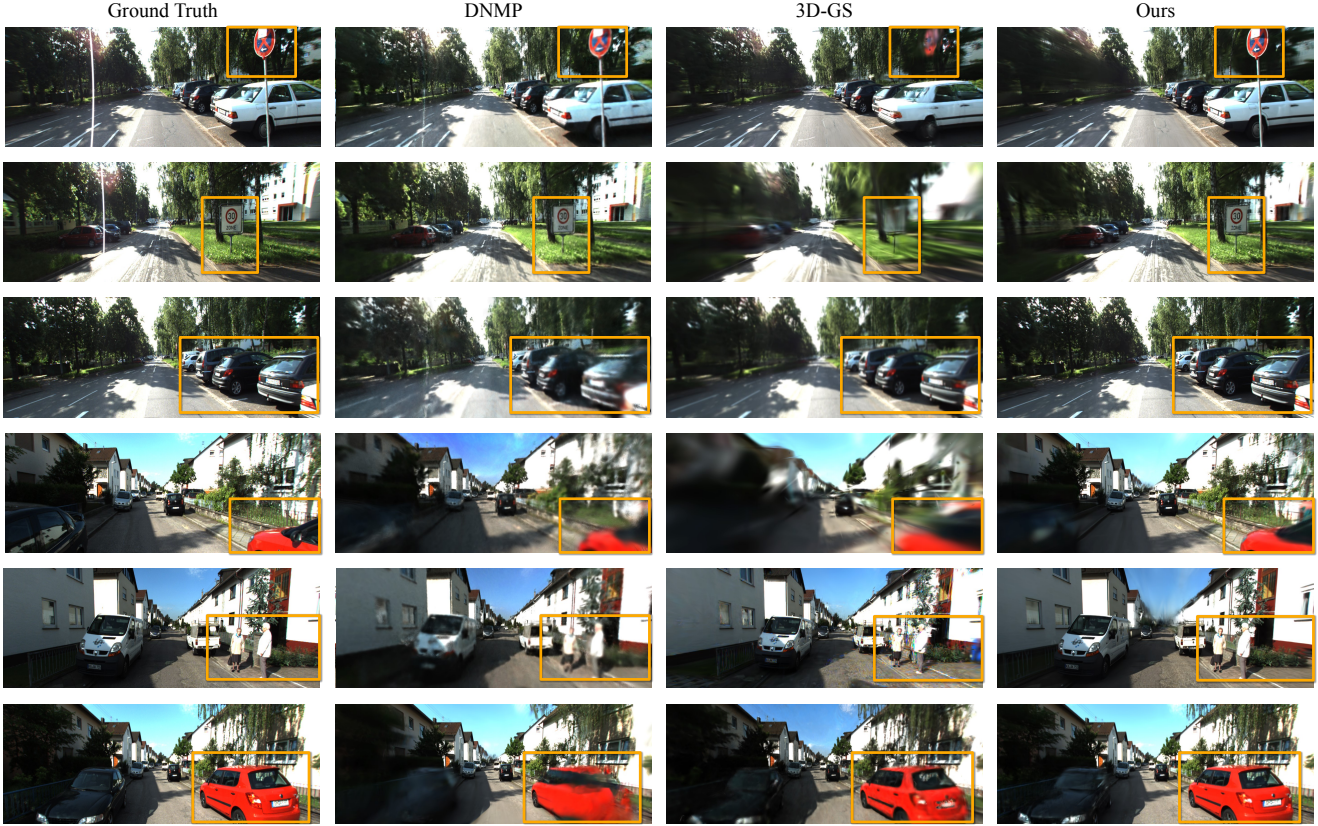}
  \caption{\textbf{Qualitative comparison on the KITTI-360 dataset.} We demonstrate the qualitative comparison results with our main competitors DNMP~\cite{lu2023urban} and 3D-GS~\cite{kerbl20233d} on driving scenes reconstruction of KITTI-360.}
  \label{360}
  \afterfig
 \end{figure*}

\noindent \textbf{Qualitative Comparison.}
We further show more qualitative results compared with the SOTA methods on the nuScenes dataset. As shown in Figure~\ref{nuScenes_comp}, our method surpasses existing works in modeling both the static background and dynamic objects in driving scenes. Please refer to the supplementary materials for video results (uploaded separately) and additional comparisons.

\section{Additional Results on KITTI-360}
\label{KITTI}
We further evaluate the performance of \ourmethod{} for monocular driving scenes on the KITTI360 dataset. We compare our method with the latest SOTA approaches trained on the KITTI360, including NeRF-based PNF~\cite{kundu2022panoptic} and 3D Gaussian-based 3D-GS~\cite{kerbl20233d}. As shown in Table~\ref{KITTI360}, our method achieves better performance than all other methods on the leaderboard.

As shown in Figure~\ref{360}, we show qualitative results compared with our main competitors on the KITTI-360 dataset. DNMP~\cite{lu2023urban} is a NeRF-based method designed for monocular driving scenes with deformable neural mesh and LiDAR prior. Our approach shows more realistic reconstruction results and fine geometry on challenging areas such as traffic signs, vehicles, people, etc. We also find that our baseline method, 3D-GS~\cite{kerbl20233d}, fails in modeling the detail areas, producing unpleasant artifacts, blurring, and unnatural colors. In contrast, although our method is not specifically designed for monocular scenarios, it still shows good adaptability and robustness in representing monocular driving scenes with detail areas and outperforms existing SOTA approaches.

\begin{table}[!t]
  \caption{
  {\textbf{Overall perforamcne of \ourmethod{} with existing state-of-the-art approaches on the KITTI-360 dataset.} We only use sequential images from a single camera as input for modeling driving scenes in the KITTI-360.}}
  \centering
  \footnotesize
  \setlength{\tabcolsep}{5mm}{
    \begin{tabular}{cccc}
    \toprule
    \textbf{Methods}  & \textbf{PSNR $\uparrow$} & \textbf{SSIM $\uparrow$} \\
    \hline 
    NeRF~\cite{mildenhall2021nerf}        & 21.94 & 0.781  \\
    Point-NeRF~\cite{xu2022point}         & 21.54 & 0.793  \\
    NSG~\cite{ost2021neural}              & 22.89 & 0.836   \\
    Mip-NeRF360~\cite{barron2022mip}      & 23.27 & 0.836  \\
    PNF~\cite{kundu2022panoptic}          & 23.06 & 0.839  \\
    SUDS~\cite{turki2023suds}             & 23.30 & 0.844  \\
    DNMP~\cite{lu2023urban}               & \cellcolor{tabthird} 23.41 & 0.846  \\
    \hline
    3D-GS~\cite{kerbl20233d}              & 22.93 & \cellcolor{tabthird} 0.847  \\
    \hline
    Ours-S                                & \cellcolor{tabsecond} 25.18 & \cellcolor{tabsecond} 0.862   \\
    \textbf{Ours-L}                       & \cellcolor{tabfirst} 25.62  & \cellcolor{tabfirst} 0.868  \\
    \hline
    \end{tabular}%
    }
  \label{KITTI360}%
  \aftertab
\end{table}%

\section{Additional Ablation Study and Analysis}
\label{ablation}
The quantitative ablation results are presented in Table~\ref{Ablation_Study} of the main text. Furthermore, we provide additional qualitative ablation comparisons to demonstrate the effectiveness of each module in our method.

\begin{figure}[!t]
  \centering
  \includegraphics[width=0.9\linewidth]{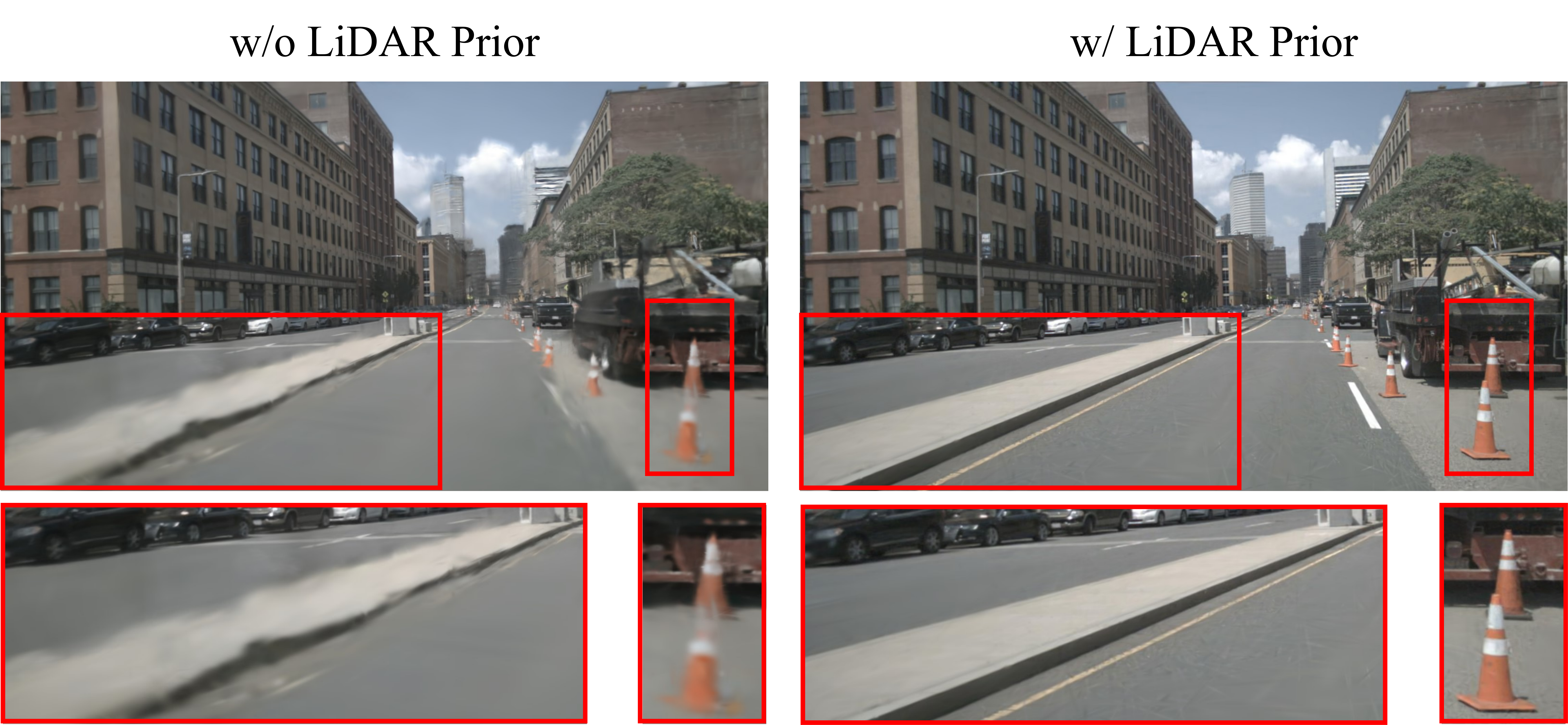}
  \caption{\textbf{Qualitative comparison with different initialization methods for 3D Gaussians.} The LiDAR prior for 3D Gaussians aids in obtaining better geometries and precise details.}
  \label{with lidar}
  \afterfig
 \end{figure}

\noindent \textbf{Initialization Methods.}
We demonstrate more qualitative comparisons with different initialization methods for 3D Gaussians in Figure~\ref{with lidar}. We can observe that utilizing LiDAR prior to initialization results in higher-quality geometry without causing noticeable distortion or blurring. In addition, details of small objects are neglected when using random or SfM initialization. Conversely, employing lidar prior enables accurate capture of these easily ignored details in both area and shape.

\begin{figure}[!t]
  \centering
  \includegraphics[width=0.9\linewidth]{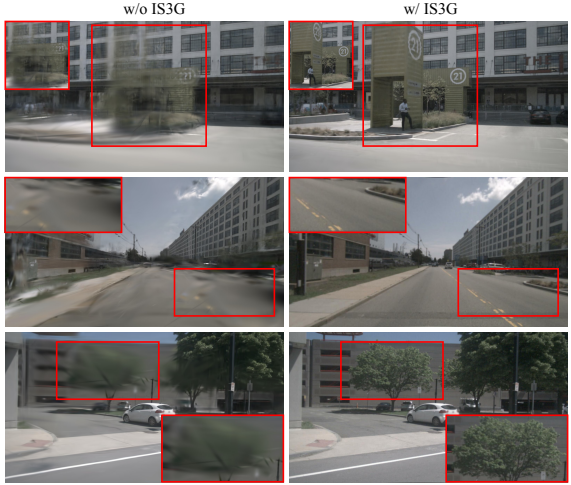}
  \caption{\textbf{Rendering with or w/o the Incremental Static 3D Gaussians (IS3G).} IS3G ensures good geometry and topological integrity for static backgrounds in large-scale driving scenes.}
  \label{static-effect}
  \afterfig
 \end{figure}

 \begin{figure}[!t]
  \centering
  \includegraphics[width=0.9\linewidth]{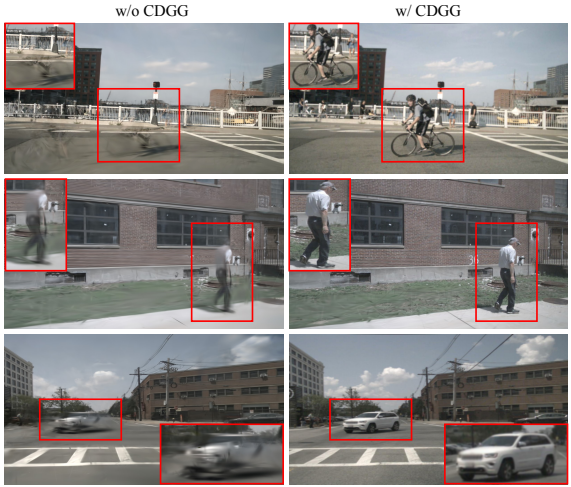}
  \caption{\textbf{Rendering with or w/o the Composite Dynamic Gaussian Graph (CDGG).} CDGG enables the reconstruction of dynamic objects at arbitrary speeds in the driving scenes (e.g., vehicles, bicycles, and pedestrians).}
  \label{dynamic-effect}
  \afterfig
 \end{figure}

\noindent \textbf{Density of Bins.}
We explore the effect of different densities of bins for reconstructing the driving scenes in Incremental Static 3D Gaussians. Here, we chose a part of the scene close to a straight line (horizontal length of about 400 meters) and cut it according to different densities of bins. The whole scene is divided into 3-7 bins, each containing multiple frames of surrounding views. As shown in Table~\ref{bin_density}, it is evident that a sparse distribution of bins results in a notable decline in performance, primarily attributable to the absence of overlapping regions among bins. Additionally, this sparse distribution may give rise to an overly extensive scale of scenes within each bin, making it impractical to adequately represent this aspect of the scene with an appropriate number of Gaussians. Alternatively, an overly dense distribution of bins may affect the Gaussian optimization efficiency between adjacent bins, leading to performance fluctuations. An appropriate distribution of bins contributes to the performance of modeling large-scale static backgrounds without wasting excessive Gaussians, thereby avoiding high computational costs.

\begin{table}[!t]
  \centering
  \caption{
  {\textbf{Effect of density of bins on the Composite Gaussian model.} N denotes for number of bins in a certain driving scene.}}
    \footnotesize
  \centering
  \setlength{\tabcolsep}{4.5mm}{
  {
    \begin{tabular}{cccc}
   \toprule
    \textbf{Bins} & \textbf{PSNR $\uparrow$} & \textbf{SSIM $\uparrow$} & \textbf{LPIPS $\downarrow$} \\
    \hline 
    N=3      &  27.94  &  0.849   & 0.256 \\
    N=4      &  28.38  &  \cellcolor{tabthird} 0.857  & 0.249  \\
    N=5       &  \cellcolor{tabthird} 28.65   &   \cellcolor{tabfirst} 0.861   &  \cellcolor{tabthird} 0.243 \\
    N=6     &  \cellcolor{tabfirst} 28.72  & \cellcolor{tabfirst}  0.861  & \cellcolor{tabfirst} 0.239 \\
    N=7     &  \cellcolor{tabsecond} 28.69  &  \cellcolor{tabsecond} 0.860  & \cellcolor{tabsecond} 0.242 \\
    \hline
    \end{tabular}%
    }
    }
  \label{bin_density}%
  \vspace{-2pt}
  \aftertab
\end{table}%

\noindent \textbf{The effectiveness of the Incremental Static 3D Gaussians.} As shown in Figure~\ref{static-effect}, the Incremental Static 3D Gaussians ensure improved geometric structure and topological integrity for the static background in driving scenes. Undesirable visual effects such as blurring, artifact, and distortion have been eliminated in the incremental reconstruction process. Due to the displacement of the ego vehicle, IS3G also ensures a good consistency of the static background captured during the ego vehicle's movement.

\begin{figure}[!t]
  \centering
  \includegraphics[width=0.9\linewidth]{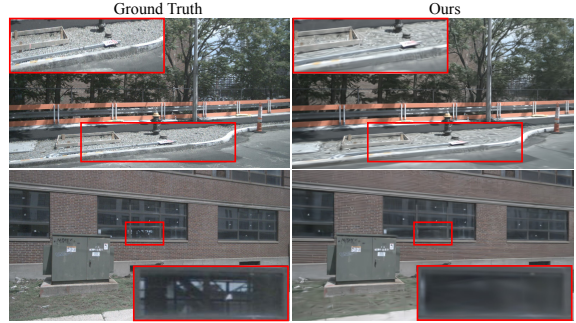}
  \caption{\textbf{Failure Cases.} Distortions exist in small objects and reflective materials (e.g., roadside pebbles and glass surfaces). }
  \label{failure}
  \vspace{-6pt}
  \afterfig
 \end{figure}

\noindent \textbf{The effectiveness of the Composite Dynamic Gaussian Graph.} As shown in Figure~\ref{dynamic-effect}, without the proposed Composite Dynamic Gaussian Graph, it would result in "invisible" or distorted dynamic objects, leading to low-quality rendering results. We can also observe that CDGG exhibits good robustness towards dynamic objects, whether they are relatively fast-moving objects (e.g., vehicles and bicycles) or slower pedestrians. CDGG enables the construction of multiple fast-moving dynamic objects in large-scale, long-term driving scenes.

\section{Failure Cases}
\label{failure case}
Our primary limitation lies in modeling extremely small and numerous objects (e.g., roadside stones) and materials with total reflection properties (e.g., glass mirrors and water surfaces), as shown in Figure~\ref{failure}. We suspect that the distortions are mainly due to 3D Gaussain's shortcomings in representing densely reflected light and errors in calculating the density of fully reflective surfaces. How to reconstruct these challenging regions will be a focus of our future research.

\end{document}